# Heterogeneous Visible-Thermal and Visible-Infrared Face Recognition using Unit-Class Loss and Cross-Modality Discriminator

Usman Cheema[1], Mobeen Ahmad[2], Dongil Han[3], and Seungbin Moon[4]*

1) Department of Computer Engineer, Sejong University, Seoul, Republic of Korea, usman@sju.ac.kr,
2) Department of Computer Engineering, Sejong University, Seoul, Republic of Korea, mobeen@sju.ac.kr,
3) Department of Computer Engineering, Sejong University, Seoul, Republic of Korea, dihan@sejong.ac.kr,
4) Department of Computer Engineering, Sejong University, Seoul, Republic of Korea, sbmoon@sejong.ac.kr.

*Abstract*—Visible-to-thermal face image matching is a challenging variate of cross-modality recognition. The challenge lies in the large modality gap and low correlation between visible and thermal modalities. Existing approaches employ image preprocessing, feature extraction, or common subspace projection, which are independent problems in themselves. In this paper, we propose an end-to-end framework for cross-modal face recognition. The proposed algorithm aims to learn identity-discriminative features from unprocessed facial images and identify cross-modal image pairs. A novel Unit-Class Loss is proposed for preserving identity information while discarding modality information. In addition, a Cross-Modality Discriminator block is proposed for integrating image-pair classification capability into the network. The proposed network can be used to extract modality-independent vector representations or a matching-pair classification for test images. Our cross-modality face recognition experiments on five independent databases demonstrate that the proposed method achieves marked improvement over existing state-of-the-art methods.

1. INTRODUCTION

The applications of facial recognition (FR) systems have increased exponentially with the advent of deep convolutional neural networks. Automated FR is being used in personal devices, public surveillance, access control, security, marketing, and other applications. FR rates on visible images have increased considerably in the past few years. However, there are limitations in using FR in scenarios involving extreme variations in illumination, expressions, pose, presentation attacks, and disguises [1–3]. Extra-visible imaging technologies are being adapted to overcome the limitations of visible imaging.

Infrared, thermal, and 3D imaging have been shown to offer advantages against pose, illumination, spoofing, and disguise. Infrared and thermal imagery is performed beyond the visible light spectrum and is less likely to be affected by variations in illumination. The infrared (IR) spectrum can be divided into reflective (active) and emission (passive) dominant bands. The reflective region consists of near-infrared (Nir) and short-wave infrared (SWIR) spectra with wavelength ranges between 0.7–1 μm and 1–2.5 μm, respectively. Infrared imagery in an active band requires an external IR source to illuminate the region of interest. The incident IR is reflected by the surface and captured by the camera. Infrared imagery in the emission-dominant band can be divided into mid-wave IR (3–5 μm) and long-wave IR (8–14 μm). MWIR and LWIR spectra are unaffected by visible light illumination, making them suitable for FR in dark environments. Additionally, Nir, MWIR, and LWIR have shown to be robust to changes in facial expressions, making them more robust than visible imagery [4]. Thermal sensors can detect the IR radiation emitted from an object, which is translated to surface temperature. Similarly, the heat emitted by the human body as IR radiation can be recorded by thermal sensors and stored as an intensity image. In addition to robustness to illumination, thermal imagery shows promise against disguise [5] and spoofing [6], making it suitable for security-sensitive applications of FR. Where visible images are easily recognizable by humans, IR and thermal images are more robust to illumination, expression, and disguise. Due to the high cost of thermal sensors, IR imagery is widely applied for CCTV and other security applications.

The increasing application of visible, infrared, and thermal imaging for FR applications has resulted in an abundance of heterogeneous data, where gallery and test images belong to different modalities such as visible images from social media such as gallery and infrared images from a CCTV footage. The scenario in which the modality of the probe images is different from that of the enrollment images is known as cross-modality FR, e.g., visible-to-Nir, visible-to-SWIR, visible-to-MWIR, visible-to-LWIR, visible-to-sketch, and 3D-to-2D FR. The problem of FR across modalities or spectra is also known as heterogeneous face recognition (HFR). Thermal-to-visible FR has received relatively less attention than other cross-modality domains owing to: (a) the high cost of thermal imaging devices, (b) the lack of availability of large-scale visible–thermal face datasets, and (c) the large modality gap between visible and thermal images owing to differences between thermal and visible image signatures. The large modality gap paired with the lack of availability of datasets makes visible–thermal (Vis–The) cross-modality face recognition a challenging task compared to Visible-to-Infrared (Vis–Nir) matching. Figure 1 shows the sample images for a subject in the visible, infrared, and thermal modalities. As can be observed, the Vis–Nir modality gap is significantly less prominent than the Vis–The modality gap, making Vis–The image matching a challenging task.

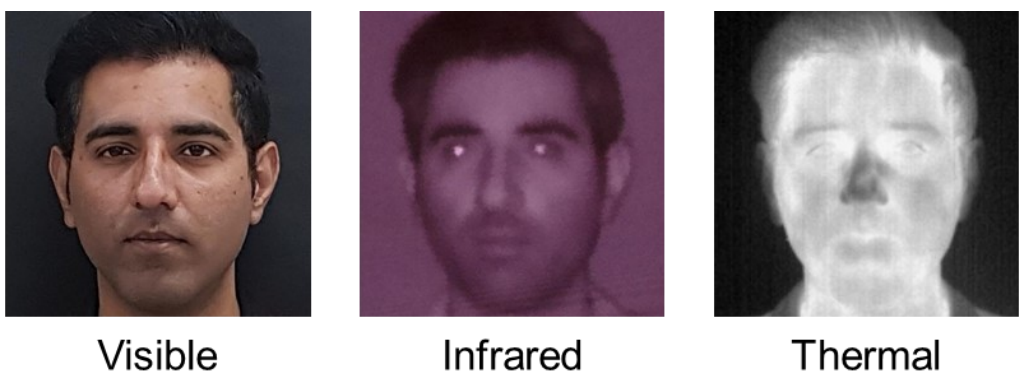

Visible         Infrared         Thermal

Fig. 1. Visible spectrum image and its corresponding infrared and thermal images.

Several studies have been devoted to reducing the cross-modality gap, e.g., synthesis-based methods, subspace projection-based methods, and local feature descriptor-based methods. However, most manually designed features are not optimized to deal with large intraclass variations and small interclass variations. Deep learning-based cross-modality face recognition techniques have been proposed, which learn deep features directly from raw data and achieve promising results in multiple cross-modality scenarios. However, existing deep learning-based approaches rely solely on sample distances or class mean distributions. Furthermore, simplistic distance metrics such as cosine or Euclidean distance are used in the succeeding step for image matching.

Motivated by the successful use of metric learning for cross-modality face recognition for Vis–Nir images, we formulate a metric learning-based loss that is robust to a small amount of training data, noisy samples, and a large modality gap. A novel Unit-Class Loss is introduced to enhance feature learning across individual samples as well as whole class distributions. We further enhance the network's same-class image pair discrimination capability by adding a distance learning module. We argue that the learned discriminator function, trained on the fly, should outperform traditional distance metrics for face image pair matching and further improve the network's ability to compute identity discriminative features.

This paper presents a novel Unit-Class descriptor and a Cross-Modality Discriminator block for visible–thermal face recognition. The proposed Unit-Class Loss can learn modality invariant identity features from raw facial images and is robust to variations in pose, expression, and disguise. In our training process, VGGFace2 [7] weights are used to initialize the backbone network for FR. The backbone network is then fine-tuned on the IRIS [8] face database with a softmax layer for visible and thermal face classification. Then, for HFR training, the proposed Cross-Modality Discriminator block is integrated with the backbone network, and the network is trained on the Vis–The database with our proposed Unit-Class Loss. The presented Unit-Class Loss and Cross-Modality Discriminator block can be integrated with any backbone network (SENet-50 [9] is used in this study). The proposed network can be used to: (a) extract modality-invariant representative features, (b) get a cross-modality imposter image pair classification, and (c) create an ensemble of embedding distances and classification probabilities. The presented algorithm is tested on multiple datasets: the TUFTS Face database [10], Collection X1 from the University of Notre Dame (UND-X1) database [11,12], University of Science and Technology China, Natural Visible and Infrared facial Expression (USTC-NVIE) database [13,14], CASIA NIR-VIS [15], and Sejong Face Database[16]. Our results show that our proposed method outperforms the existing methods for Vis–The HFR and achieves improved results over the state-of-the-art methods for Vis–Nir HFR.

In this article, our main contributions are as follows:

- A simple and efficient end-to-end mechanism for heterogeneous face recognition using a single-stream CNN backbone is proposed.

- A hybrid Unit-Class Loss ($L_{uc}$) for optimizing cross-modality face recognition is introduced. $L_{uc}$ allows the network to learn modality-independent, identity-discriminative features through penalizing the network for large intraclass variations and small interclass distances.

- We propose a Cross-Modality Discriminator block, trained online, that differentiates between same-class samples and different-class samples without the need for any hard-coded distance metric.

- We demonstrate the superior performance of our proposed methodology over previous works on various visible–thermal and visible–infrared face datasets.

The rest of the paper is organized as follows. In Section II, we review related works on Vis–Nir and Vis–The HFR. Section III presents the details of the proposed methodology. The results on HFR datasets are presented in Section IV and Section V presents our conclusions.

## 2. RELATED WORKS

Heterogeneous matching across different imaging modalities has received increasing attention from the biometrics community. Visible-to-thermal face recognition poses a particularly challenging problem for HFR. We review the closely related Vis–Nir and Vis–The literature.

### 2.1. Near-Infrared-to-Visible Face Recognition

Various approaches have been proposed to solve the problem of heterogeneous face recognition. The most common approach is to derive a set of features from both spectra, such that the modality information is removed, and the identity discriminative features are retained. This is done by finding a mapping for both modalities to a common feature subspace. There has been a significant improvement in Vis–Nir [17,18] and SWIR–Vis matching accuracy [19,20] in recent times. Some of the initial work in Vis–Nir matching is based on shared representation learning. Yi et al. [21] extracted Gabor features at localized facial points and then used Restricted Boltzmann Machines (RBMs) to remove the heterogeneity around the focal points by learning a shared representation. Then, these shared representations of local RBMs were connected and processed using PCA. Zhu et al. [22] proposed a transductive heterogeneous face-matching method, which adapts the Vis–Nir matching learned from available training pairs to the subjects in the target set. Ghosh et al. [23] proposed a subclass heterogeneity-aware loss to minimize the intraclass distance and maximize the interclass distance across the modalities. Nir/SWIR imaging can help build a robust face recognition system. However, it has limited utility in uncontrolled environments with disguise and spoofing. Therefore, in recent times, HFR focus has shifted towards visible-to-thermal image matching [24–26].

### 2.2. Thermal-to-Visible Face Recognition

Algorithms for visible-to-thermal image matching can be categorized as synthesis-based methods or local feature-based methods. Synthesis-based methods aim to generate cross-modality images via image synthesis and try to solve the problem of modality difference during a preprocessing stage [27]. These methods learn a mapping from probe modality to the enrollment modality. Synthesis-based HFR is a two-step process; first, it tries to synthesize a visible image from a thermal image and then applies face recognition to match the synthesized image with its gallery pair. This simplifies the HFR problem to a simple homogeneous face recognition problem. Li et al. [28] is one of the earlier attempts to synthesize visible images from thermal counterparts. In recent times, deep-learning-based image synthesis methods have been applied to HFR. Iranmesh et al. [27]

proposed to use a coupled GAN consisting of two generators and two discriminators to synthesize visible images from thermal images.

As machine learning algorithms are not 100% accurate, a multistep HFR process leads to error accumulation at each step. Furthermore, a larger hyperparameter space needs to be explored to find the optimal solution, which is a time-consuming task for large networks. Consequentially, the cost of training is increased when training multiple systems. No mathematical mapping exists between visible and thermal faces. The visible image is captured based on the visible light reflected from the facial surface and depends on the face structure. The thermal image is captured based on the heat radiated from the face, which depends upon the underlying vascular structure, body temperature, and various other factors [29]. GAN methods, at best, can generate example-based thermal faces rather than finding a visible-to-thermal mapping.

Single-stage approaches such as local feature-based methods can be easier to optimize and redeploy for various scenarios. In local feature-based methods, features are extracted from a thermal probe image and compared to the features extracted from visible images in the gallery [30]. Bourlai et al. [31] fused textural descriptors to improve the recognition performance across modalities. Klare et al. [24] proposed a generic HFR framework where the probe and enrollment images are represented in terms of nonlinear similarity by using a prototype random subspace (similarity kernel space) such that the prototype subjects each have an image in both modalities. LWIR images are of considerably low resolution because of long-wave sensing; hence, they are challenging to match to visible images. Owing to the higher emissivity sensing range of LWIR, they are well suited for environments with an absence of light. Some studies have been conducted in the LWIR-to-visible matching domain. Choi et al. [25] used partial least-squares-discriminant analysis (PLS-DA) based techniques to reduce the modality gap. Hu et al. [26] proposed to use discriminant partial least squares (PLS) by specifically building PLS gallery models for each subject with the help of thermal cross examples. Eslam et al. [32] proposed to recognize facial features in thermal images by employing Haar features and AdaBoost; however, they used visible as well as thermal images in the gallery. Reale et al. [33] applied coupled dictionary learning to the thermal-to-visible matching problem. Riggan et al. [34] used coupled auto-associative neural networks (CpNN) along with deep perceptual mapping (DPM) to learn common latent features that are useful for cross-modal face recognition.

Various deep learning approaches have been proposed for cross-modality face recognition. Triplet Loss and its variations have been used for face verification, but these techniques are sensitive to training images and require hard-sample mining for effective training. Synthesis-based methods rely on effective training of generative adversarial networks, which is an inherently a complex problem. Facial feature-based methods depend on facial alignment, which is an independent problem for highly noncorrelated modalities. Various effective approaches for Vis–Nir have been proposed using multiple network branches, which increases network parameters multifold and requires large training resources. Visible–Thermal face recognition faces the challenges of a large modality gap, low modality correlation, and lack of large databases. We propose a single-stream algorithm for Visible–Thermal heterogeneous face recognition.

## 3. Proposed Method

This section presents the building blocks of our proposed algorithm. The goal of an efficient cross-modality recognition system is to project the input images such that the intraclass projections have a small distance whereas interclass projections have a large distance. The choice of our backbone architecture, the motivation for our loss function, and the proposed loss function and discriminator block for achieving this goal are explained below.

### 3.1. Backbone Architecture

Deep residual networks were introduced to solve the issue of degradation in deep neural networks [35]. They introduced skip connections with the identity function to allow the gradient of the cost function to progress directly from deeper layers to the surface layers. the identity blocks are also effective at improving the performance of networks when vanishing gradients and degradation are not an issue. Similar skip connection approaches have been used for cross-modality face recognition [27,36–38]. ResNet-50 [35] has been successfully used for face recognition with high accuracy on the MS1M [39] and VGGFace2 [7] datasets. Squeeze and Excitation block (SE) blocks recalibrate channel-wise feature responses by explicitly modeling the interdependencies between channels. The SE block models channel interdependencies by selectively emphasizing informative channels and suppressing less useful ones. For the transformation $x_l$ to $x_{l+1}$, squeeze ($F_{sq}$), excitation ($F_{ex}$), and scaling ($F_{sc}$) operations are performed for a given feature matrix $x_l$ of size H × W × C. The advantages of integrating SE blocks have been demonstrated for object and scene classification during ILSVRC 2017 and face recognition using VGGFace2. We use the SENet-50 architecture as our backbone network. Our network takes a 224 × 224-sized image input and calculates a 256-dimensional vector representation of the input image.

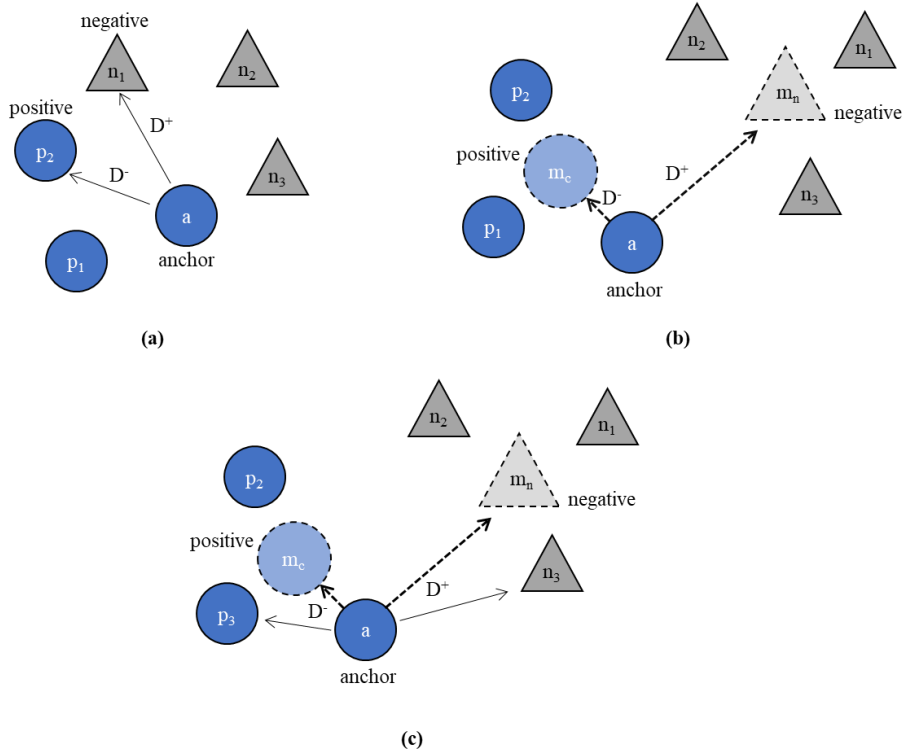

Fig. 2. (a) Triplet loss, (b) Class mean triplet loss, (c) Unit-Class Loss.

### 3.2. Triplet Loss

Triplet Loss was used by Schroff et al. [40] for face recognition and clustering. Since then, triplet loss and its variations have been used for single-modal, multi-modal, and cross-modal face recognition. For the calculated vector representations of a given anchor image, positive image, and negative image, the triplet loss aims to minimize anchor-to-positive distance and increase anchor-to-negative distance. The idea is to train the network such that the same-class images are projected in the nearby region, whereas off-class images are projected farther from the anchor.

$$L_T(a, p, n) = Max(0, D(a, p) - D(a, n) + \alpha) \qquad (1)$$

$$a = \Theta(a),$$

where $\Theta(\cdot)$ is the feature extraction function of the network, a is the anchor image, and $L_T(a, p, n)$ is the loss function for the vector representations of $a$, positive class image $p$, and negative class image $n$. $D(\cdot, \cdot)$ is the distance function for the learned vector representations of two images. L2 or cosine distances are typically used as distance measures. A graphical illustration of triplet loss is shown in Fig. 2 (a).

### 3.3. Class Mean Triplet Loss

The goal of a heterogeneous face recognition system is twofold: (a) to minimize the distances between multimodal image representations of the same class and (b) to increase the distance between image representations of different classes. Given a vector representation of an image of class $c$, the class mean triplet loss is defined as follows:

$$L_S\,(a_c, m_c, m_n) = Max(0, D(a_c, m_c) - D(a_c, m_n) + \alpha). \qquad (2)$$

Here,

$$m_c = \frac{1}{l_c} \sum_{k=1}^{l_c} \Theta(x_k),$$

$$m_n = \frac{1}{l_n} \sum_{k=1}^{l_n} \Theta(x_k), \text{ and}$$

$$n \in \{U - c\}.$$

where, $U$ is the universal set containing the classes. $L_S\,(a_c, m_c, m_n)$ is the loss function for the vector representations of the anchor image $a_c$, belonging to class $c$. $m_c$ and $m_n$ are the means of vector representations in classes $c$ and $n$, respectively. Figure 2(b) shows the conceptual representation of the class mean triplet loss.

*3.4. Unit-Class Loss*

The training procedure for triplet loss is sensitive to the selection of effective samples, noise, outliers, number of classes, and minibatch diversity. Using class means for loss calculation has been shown to be more robust to noise and random sample selection. However, convergence becomes more difficult because of the larger number of parameters involved. We present a novel Unit-Class Loss ($L_{UC}$) that combines principals from triplet loss and class mean triplet loss to benefit from the advantages of both, as shown in Fig. 2 (c). The weight parameter $\beta$ is introduced for the optimal weight distribution of sample-based and class-mean-based optimization of gradients. The values for $\alpha$ and $\beta$ are determined empirically.

$$L_{UC} = Max(0, (1 - \beta)\,(\,D(a_c, m_c) - D(a_c, m_n)) + \beta\,(\,D(a, p) - D(a, n)\,) + \alpha\,) \qquad (3)$$

*3.5. Cross-Modality Discriminator Block*

We hypothesize that a network trained with the end goal of matching cross-modal face pairs would train more efficiently than a network trained to project class vector representations. Traditionally, vector distance measures such as Euclidean or cosine distance have been employed to calculate the similarity between the vector representations of two images. We introduce a Cross-Modality Discriminator (CMD) block to distinguish same-class face images from different class face images.

An *L2*-normalized 256-dimension vector representation of the input image is obtained from the backbone network and processed such that each image embedding is concatenated with every image. The $b \times 256$ output, where b is the batch size and 256 is the representation vector dimension, is remapped to the $b \times b \times 512$ ($batch\ size \times batch\ size \times vector\ size$) dimension input for the CMD block.

$$CMD\ input = a_i || a_{j,} \text{ where } i, j \in U$$

where, $a_i$ and $a_j$ are the vector representations of images $i$ and $j$, respectively, while $b$ is the total number of images in the batch. The goal is to train the CMD block to classify the same-class images from concatenated image embeddings of an image pair. These concatenated image pair representations are then fed into subsequent dense layers followed by the sigmoid activation layer, as shown in Fig. 3. The proposed CMD block is trained using binary cross-entropy loss to classify the matching image pairs. Our experiments demonstrate that the proposed block helps train the network for improved performance and outperforms traditional distance measures for face matching with a small increase in computational cost.

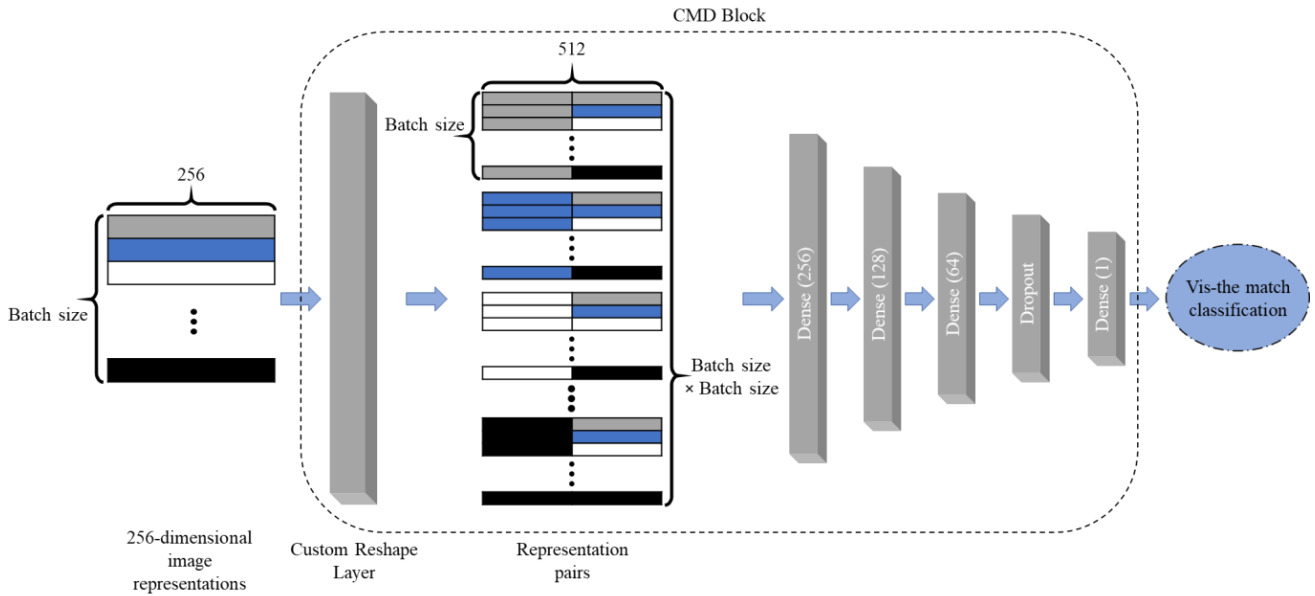

Fig. 3. Cross-Modality Discriminator (CMD) block architecture.

### 3.6. Training Process

This section presents the proposed network and our training algorithm in detail. We aim to reduce intraclass variation across the visible and thermal domains while increasing interclass distances. Furthermore, the proposed CMD block trains the network to differentiate between same-class and different-class facial images. The proposed architecture preparation steps are simple and can be integrated with the existing state-of-the-art architectures. Figure 4 shows an overview of the proposed methodology.

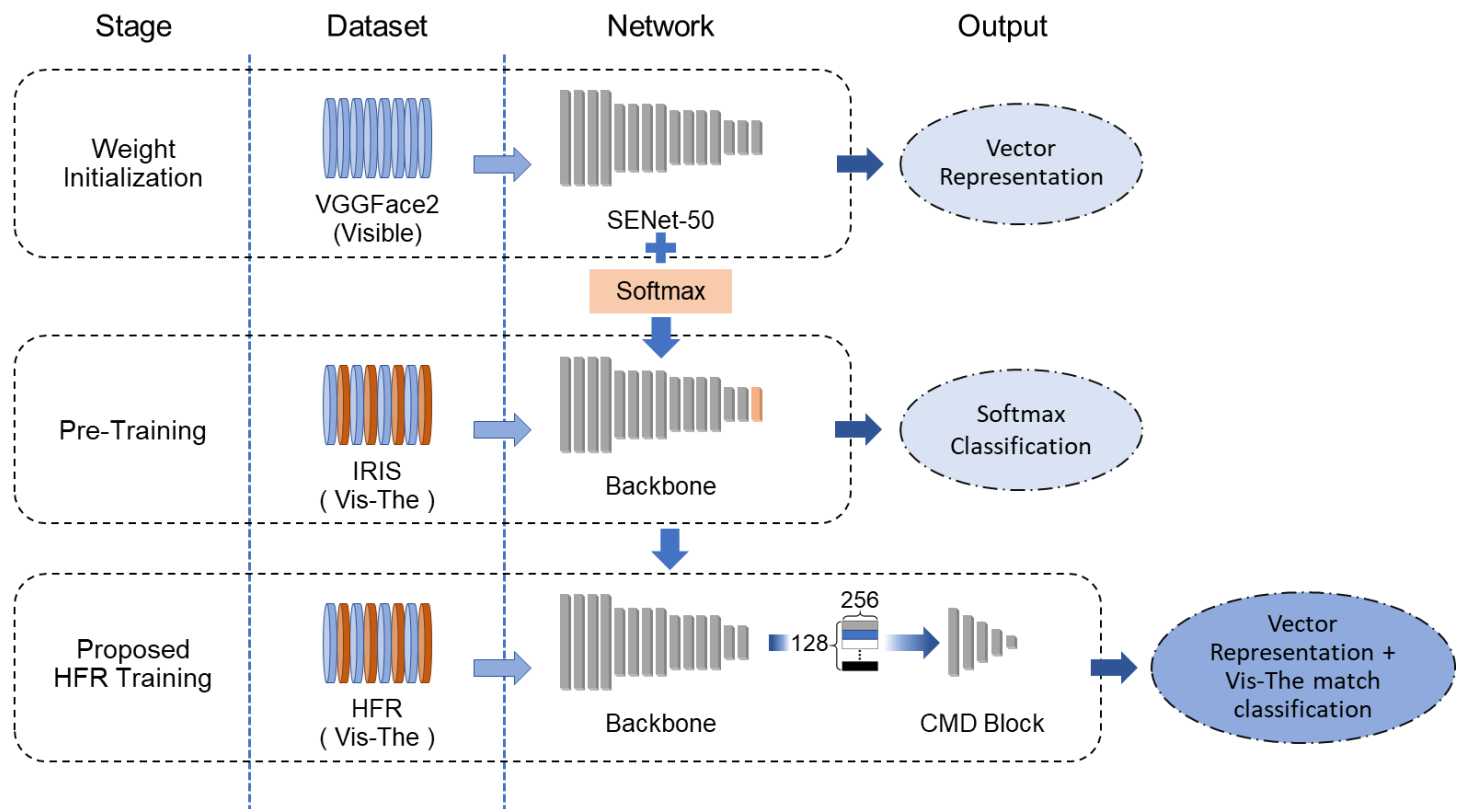

Fig. 4. Overview of the proposed methodology and training pipeline.

Our training process consists of three stages: weight initialization, pretraining, and HFR training. Network weights from a SENet-50 trained on the VGGFace2 [7] face database are used to initialize the backbone network. As the VGGFace2 dataset contains only visible images, we pretrain the base network on the Vis–The face dataset to learn thermal as well as visible feature extraction. The IRIS [8] face dataset is used to pretrain the network using cross-entropy loss. The dataset contains 2,552 images each in visible and thermal modalities for 29 subjects. The facial images contain variations in expression, pose, and illumination. Feature-wise centering and feature-wise standard normalization calculated on the entire dataset (visible and thermal images) are applied to each sample. Furthermore, geometric transformations are used as data augmentation techniques to mitigate overtraining. The training data is fed in a pseudo-random manner, ensuring a randomized class and modality distribution. The network is trained until there is no further decrease in loss. We choose cross-entropy loss to train the network at this stage, given its established performance and stable nature for FR algorithms.

For cross-modality face recognition, the last softmax layer from the pretraining stage is replaced with an L2-normalized dense layer and the network is connected to the proposed CMD block. The resulting network contains two outputs, i.e., the L2-normalized vector representations and the binary classification from the CMD block. We optimize the network using the proposed Unit-Class Loss and binary cross-entropy for vector representations and CMD output, respectively. Class identities are passed as label information to Unit-Class Loss and the CMD block. The pretraining of the backbone network and the training process of the proposed network are summarized in Algorithm 1 and Algorithm 2, respectively. It should be noted that our training and testing procedure does not require specific image preprocessing, facial alignment, or landmark labeling for training or testing.

---

**Algorithm 1** Pretraining the backbone network

---

**Require:** VGG2Face weights, IRIS dataset.

**Ensure:** Network parameters $\Theta$.

1. Load the VGGFace2 weights

2. **For** $t = 1, \dots, T$ **do**

   a. CNN optimization on IRIS dataset

   b. Update $\Theta$ via back-propagation

3. **Return** $\Theta$;

---

---

**Algorithm 2** Training the Proposed Network on HFR data

---

**Require:** Pretrained backbone network weight $\Theta$, HFR dataset, and loss weights $\mu$ and $\beta$.

**Ensure:** Network parameters $\Theta_{\text{HFR}}$.

1. Load the pretrained weights

2. Replace softmax layer

3. Add discriminator block

4. **For** $t = 1, \dots, T$ **do**

   a. CNN optimization on HFR dataset

   b. Update $\Theta_{\text{HFR}}$ via back-propagation using $L_{UC}$ and $L_{CMD}$

6. **Return** $\Theta_{\text{HFR}}$;

---

## 4. EXPERIMENTAL RESULTS

In this section, we evaluate the performance of the proposed architecture on popular Vis–The face datasets. We provide comprehensive parameter settings and implementation details for research reproducibility. Rank-1 accuracy and verification rates (VR) at False Acceptance Rate (FAR) of 1% and 0.1% are compared to previous HFR results, where available.

*4.1. Databases*

The experiments were performed on USTC-NVIE [13,14], TUFTS [10], UND-X1 [11,12], and Sejong Face Database [16]. To validate the consistency of the proposed algorithm, we also include our results on CASIA NIR-VIS 2.0 face database [15], a visible and near-infrared face image dataset. Figure 5 shows the sample images from the HFR database.

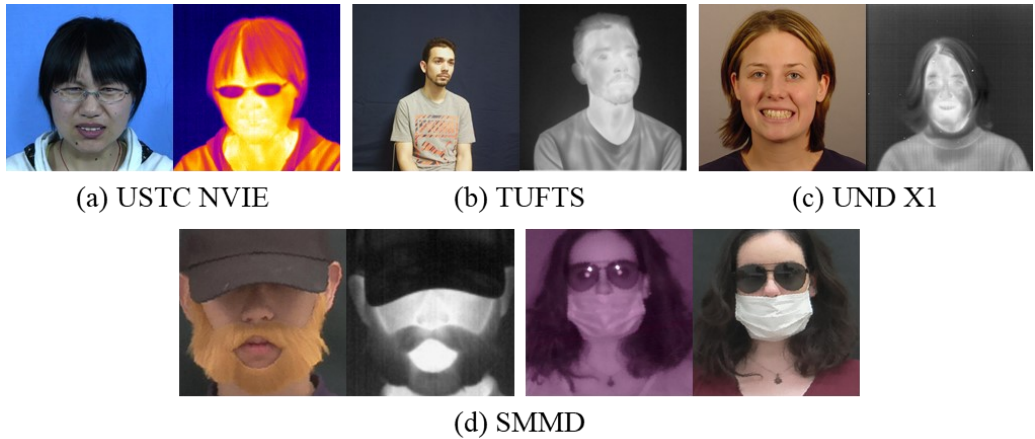

(a) USTC NVIE      (b) TUFTS      (c) UND X1

(d) SMMD

Fig. 5. Sample visible (left) and thermal (right) images from (a) USTC-NVIE, (b) TUFTS, (c) UND-X1, and (d) Sejong Multi-modal Face Database.

*4.1.1. USTC-NVIE*

The USTC-NVIE database [13,14] is a facial expression database with visible–thermal image pairs for 215 subjects. The database is further divided into two subsets: a spontaneous database consisting of image sequences from onset to the apex of facial expressions and a posed database consisting of apex images. The images are captured with three illumination variations, i.e., illumination from left, right, and front. This is an expression recognition database but is used for Vis–The HFR in this study. Among the 215 subjects, 126 subjects were found to have usable data (having sufficient images in both modalities). Both subdatabases are used to maximize the number of training and test data. The training was performed using 32 image pairs (visible and thermal images) for 100 subjects. Thirty-two image pairs and 1600 images for 26 subjects were used for testing.

*4.1.2. TUFTS*

The TUFTS database [10] contains data belonging to multiple categories, i.e., 2D visible, thermal, Infrared, 3D, 3D LYTRO, Sketch, and Video. The database contains images having variations in pose, expression, and sunglasses. To avoid the additional challenge of pose, only frontal images with variations in expression and glasses were used for training and testing. Thermal and visible image corpus was used for our HFR experiments. The training was performed using 5–8 image pairs (visible and thermal images) for 74 subjects. Here, 5–8 image pairs and 500 images for 38 subjects were used for testing.

*4.1.3. UND-X1*

UND-X1 [11,12] contains 82 subjects with LWIR and visible light image pairs with varying illumination, expression, and time-lapse. Out of 82 subjects, 50 subjects' images were used for training. Forty image pairs were used for each subject. A single gallery image (visible) and 40 probe images (thermal) from 32 subjects were used for FR. Verification experiments were performed on 1280 images for 32 test subjects.

### 4.1.4. Sejong Face Database

Sejong Face Database [16] (SFD) is a proprietary face dataset that will be publicly available to the research community after its publication. The database contains images in visible, infrared, and thermal modalities for 100 subjects. The subjects are captured wearing disguises and facial add-ons such as fake beards, caps, scarves, and wigs, with and without makeup, and so on. The addition of facial add-ons makes HFR a more challenging problem for SFD. Forty image pairs from 75 subjects were used for training and an equal number of image pairs for 25 subjects were used for testing.

### 4.1.5. CASIA NIR-VIS 2.0

The CASIA NIR-VIS 2.0 [15] database contains visible–infrared image pairs for 725 subjects. The number of image pairs varies per subject. The number of images per person ranges between 1–22 and 5–50 for visible and infrared, respectively. The database contains two views of the evaluation protocols. View-1 is used for training and View-2 is used for testing. For a fair comparison with other results, we follow the standard protocol in View-2 for testing.

### 4.2. Training Parameters and Implementation Details

The proposed network is implemented using TensorFlow [41], an open-source deep learning framework. The experiments are performed using two Nvidia GTX 1080Ti GPUs. In the pretraining stage, we adopt the IRIS Visible–Thermal image dataset to train the backbone network. At this stage, the categorical cross-entropy loss is used for multimodal face recognition. Here, 5,100 images for 29 subjects are used as training data. The training images are resized to a size of $224 \times 224$ pixels and the labels contain class information only. The network is trained using an Adam optimizer [42] with an initial learning rate of $3\mathrm{e}^{-4}$ and the learning rate is reduced by a factor of 0.8 when the error plateaus. The network is trained using a mini-batch size of 64 (32 visible and 32 thermal images) until the loss plateaus.

In the training stage, the softmax layer is replaced with a 256-dimension $L$2-normalized dense layer and the CMD block is added to the head of the network. The backbone network is trained on the HFR dataset with the proposed Unit-Class Loss and the CMD block is trained using cross-entropy loss. The values of $\alpha$ and $\beta$ are set to 1.6 and 0.6, respectively. The batch size is set to 128 (64 visible and 64 thermal) images. The Adam optimizer, with an initial learning rate of $3\mathrm{e}^{-4}$, is used for gradient descent optimization. The learning rate is reduced by a factor of 0.8 when the loss plateaus. The fine-tuning process for each dataset takes approximately 2 hours.

Owing to the difference in image sizes between different modalities and across different databases, the images are cropped to match the shorter edge. The square images are then scaled to $224 \times 224$ pixels for all databases. Data augmentation is performed to mitigate overtraining and increase training size. Geometric transformations for rotation, shift in both axes, brightness shift, shear, and horizontal flip are applied to the training data. Image normalization is performed using feature-wise centering and feature-wise standard normalization calculated on the entire training dataset.

### 4.3. Training and Test Data

To verify the performance of our proposed HFR algorithm, we compare our method with the state-of-the-art HFR methods. The number of available test subjects, gallery images, and probe images for the face verification experiments is listed in Table 1.

Table 1. The number of available test subjects, training images and test images for each subject.

| Database | Subjects | Train Images / Subject | Test Images / Subject |
|---|---|---|---|
| UND-X1 | 32 | 40 | 40 |
| TUFTS | 38 | 8 | 8 |
| NVIE | 26 | 32 | 32 |
| SFD | 30 | 70 | 70 |

The proposed network can be used in multiple ways: (a) the *L2*-normalized features (Embd) can be retrieved from the *L2*-normalized layer for a single image and HFR performed using the cosine distance between gallery and probe images, (b) a binary classification, i.e., genuine versus imposter image pair, for an image pair can be extracted from the CMD block output, (c) score fusion (Fusion) for distance measure of embeddings and CMD block can be performed to achieve an average recognition measure. The Rank-1 accuracy and VR for Embd, CDM, and Fusion are presented. The tests are performed using one visible image against all available thermal images as gallery and probe images, respectively. For testing the CMD classification, all available Vis–The image pairs from the test set are used. Fusion score is reported on the test pairs used for embedding recognition.

### 4.4. Evaluation Metrics

Given two test images, the 1:1 verification determines if the images belong to the same identity. For the embedding output of the model, the distances are calculated using the cosine distance between the extracted embedding features and the same or different identity classification performed based on a threshold value. For the CMD output of the model, the model outputs the same identity probability for the input image pair. $1:N$ Identification determines the matching identity from a gallery of N images for a given probe image. Rank-1 matching accuracy is reported for the embedding output of the proposed model.

We present Rank-1 recognition rate, True Positive Rate (TPR) at False Acceptance Rate (FAR) of 1% and 0.1%. Rank-1 accuracy is defined as the proportion of correctly predicted face image pairs (True positive) among the total number of image pairs.

$$Accuracy = \frac{TP + TN}{TP + TN + FP + FN}$$

Where TP = True positive, FP = False Positive, FP = False positive, and FN = False Negative.

TPR at an FAR is described as the number of correct predictions at a specific percentage of acceptable false predictions. TPR and FAR are calculated as

$$TPR = \frac{TP}{(TP + FN)}$$

$$FAR = \frac{FP}{(FP + TN)}$$

A threshold $t$ is defined to determine the FAR. FMR and FAR are interchangeable and calculated the same way.

### 4.5. Qualitative Results

Table 2 presents the Rank-1 identification and verification rates on UND-X1. In this experiment, 1280 and 51,000 Vis–The image pairs are used for embedding and CMD classification testing, respectively. We compare the performance of our method with those of the recent CpGAN [27] and DPM [43]. The proposed network achieves higher Rank-1 accuracy and verification metrics on the UND-X1 dataset compared to other methods, as listed in Table 2.

Table 2. Rank-1 accuracy and verification rate on UND-X1 database.

| Method | Rank-1(%) | TPR@FAR = 1% | TPR@FAR = 0.1% | Dim |
|---|---|---|---|---|
| DPM [43] | 83.73 | | | |
| CpGAN [27] | 76.4 | - | - | - |
| Proposed Embd | 80.8 | **90.8** | 52.9 | 256 |
| Proposed CMD | 93.6 | 87.7 | **64.8** | 1 |
| Proposed Fusion | **95.2** | 88.3 | 61.1 | - |

We compare our results with the existing results on the TUFTS dataset. As the TUFTS database is relatively new, few Vis–The HFR results have been reported. In our experiment, 304 and 2432 Vis–The image pairs are used for embedding and CMD classification testing, respectively. DVG-Face [44] proposes a dual variation generation method to generate thermal images from visible images. The generated dataset is used to train a LightCNN for recognition. While we achieve a marginally improved Rank-1 recognition rate using embedding distances, the proposed CMD classification block and ensemble achieve significant improvement over the current methods, as presented in Table 3. Baseline results using triplet loss for the TUFTS database, are presented in ablation studies.

Table 3. Rank-1 accuracy and verification rate on TUFTS database.

| Method | Rank-1(%) | TPR@FAR = 1% | TPR@FAR = 0.1% | Dim |
|---|---|---|---|---|
| DVG-Face [44] | 75.7 | 68.5 | 36.5 | - |
| Proposed Embd | 79.2 | 94.7 | 45.6 | 256 |
| Proposed CMD | 97.0 | **95.7** | 48.0 | 1 |
| Proposed Fusion | **98.5** | 95.3 | **49.1** | - |

As USTC-NVIE is primarily an expression database, it lacks significant HFR results. We report baseline results for triplet loss in ablation studies. Here, 832 and 26,600 Vis–The image pairs are used for embedding and CMD classification testing, respectively.

The PCA, Fisherface, and GR-HFR results are presented from a previous study [45]. Our proposed method achieves higher recognition and verification metrics than existing methods as can be seen in Table 4.

Table 4. Rank-1 accuracy and verification rate on USTC-NVIE database.

| Method | Rank-1(%) | TPR@FAR = 1% | TPR@FAR = 0.1% | Dim |
|---|---|---|---|---|
| PCA | 0 | - | - | - |
| Fisherface | 8.72 | - | - | - |
| GR-HFR | 77.4 | - | - | - |
| Proposed Embd | 99.4 | **98.4** | **71.0** | 256 |
| Proposed CMD | 99.3 | 96.5 | 69.9 | 1 |
| Proposed Fusion | **99.7** | 97.4 | 70.3 | - |

The Sejong Face Database has been proposed for disguised face recognition across various modalities. The inclusion of facial add-ons that hide large parts of the face makes it a particularly challenging face dataset. Here, 2,100 and 147,000 Vis–The image pairs are used for embedding and CMD classification, respectively. Owing to the lack of HFR results on SFD, we include face recognition rates for (a) single-modal visible images and (b) multimodal images (visible, thermal, and infrared) using score fusion on the database as reported in the database paper. Furthermore, cross-modality recognition rates using triplet loss are reported. Our proposed method achieves recognition and verification metrics that are comparable to single-modal face recognition methods and performs better than the baseline methods as shown in Table 5.

Table 5. Rank-1 accuracy and verification rate on Sejong Face database.

| Method | Rank-1(%) | TPR@FAR = 1% | TPR@FAR = 0.1% | Dim |
|---|---|---|---|---|
| Usman (a) | 72.1 | - | - | - |
| Usman (b) | 90.3 | - | - | - |
| Triplet Loss | 60.4 | - | - | - |
| Proposed Embd | 73.2 | 74.5 | 32.4 | 256 |
| Proposed CMD | 91.6 | **87.7** | **41.9** | 1 |
| Proposed Ens | **92.4** | 85.4 | 40.9 | - |

CASIA NIR-VIS 2.0 is used to present our results on Vis–Nir modality. For fair comparison, we perform the same set of experiments as in a previous study [46]. The dataset has two subsets; View-1 is used for training and View-2, with 10 different splits, is used for testing. The number of images in the dataset is approximately 1000 visible images and 1500 infrared images during the testing phase. The CASIA NIR-VIS 2.0 restricts the testing to one gallery image per subject. Therefore, there are 358 visible gallery images and about 6000 probe infrared images for testing. Table 6. summarizes the results of our proposed model compared to other recent methods applied on CASIA NIR-VIS 2.0. It should be noted that the proposed network is designed and optimized to perform on Vis–The data, yet we achieve improved results on the CASIA NIR-VIS database.

Table 6. Rank-1 accuracy and verification rate on CASIA NIR-VIS database.

| Method | Rank-1(%) | TPR@FAR = 1% | TPR@FAR = 0.1% | Dim |
|--------|-----------|--------------|----------------|-----|
| IDNet [47] | 87.1 ± 0.9 | - | 74.5 | 320 |
| IDR [48] | 97.3 ± 0.4 | 98.9 ± 0.3 | 95.7 ± 0.7 | 128 |
| W-CNN [49] | 98.7 ± 0.3 | 99.5 ± 0.1 | 98.4 ± 0.4 | 128 |
| Proposed Embd | 97.5 | **99.8** | 78.2 | 256 |
| Proposed CMD | 99.2 | 99.5 | 78.6 | - |
| Proposed Ens | **99.5** | 99.3 | **77.8** | |

## 5. ABLATION STUDIES

To verify the effectiveness of our proposed method, we perform ablation studies to explore the effects of different loss functions, the CMD block, and different values of $\alpha$ and $\beta$. The Rank-1 accuracy and VR for the USTC face database are reported in Table 7. The results are calculated using embedding distances for a fair comparison. The ablation experiments are performed as follows:

1. Experiment 1: The SENet-50 backbone network is trained using triplet loss without the CMD block.

2. Experiment 2: The SENet-50 backbone network is trained using the class mean triplet loss without the CMD block.

3. Experiment 3: The network is trained using the proposed Unit-Class Loss without the CMD block.

4. Experiment 4: The network is trained using the proposed Unit-Class Loss without the CMD block.

5. Experiment 5: The network is trained using triplet loss and the CMD block is added.

6. Experiment 6: The network is trained using the proposed Unit-Class Loss and the CMD block (our proposed method).

Table 7. Rank-1 accuracy and verification rate for the ablation studies on the USTC database.

| Method | Rank-1(%) | TPR@FAR = 1% | TPR@FAR = 0.1% |
|--------|-----------|--------------|----------------|
| Exp. 1: Triplet Loss | 95.6 | 86.1 | 43.1 |
| Exp. 2: Class mean Loss | 97.8 | 91.2 | 53.8 |
| Exp. 3: Class mean Loss + CMD block | 98.4 | 93.4 | 64.2 |
| Exp. 4: Unit-Class Loss | 97.9 | 88.4 | 56.2 |
| Exp. 5: Triplet Loss + CMD block | 98.6 | 96.1 | 67.1 |
| Exp. 6: Proposed (Unit-Class Loss + CMD block) | **99.4** | **98.4** | **71.0** |

As can be seen in Table 7, the combination of Unit-Class loss and CMD block outperforms the ablation variations in terms of Rank-1 accuracy and VR. The addition of CMD block with triplet loss achieves the second-best performance as the network is reinforced by the final goal of HFR as well as class distance modulation.

The effects of changing the values of $\alpha$ and $\beta$ are shown in Table 8, as can be seen that changing the value of $\beta$ affects the network performance, whereas changing the value of $\alpha$ does not have a noticeable effect on performance. We determine the values $\alpha = 1$ and $\beta = 0.5$ to be optimal for our HFR results.

Table 8. Performance of the proposed method with different values of α and β.

| α | β | Rank-1(%) |
|---|---|---|
| 0.5 | 0.2 | 98.7 |
| 1 | 0.2 | 98.5 |
| 2 | 0.2 | 98.6 |
| 0.5 | 0.5 | 99.4 |
| 1 | 0.5 | **99.4** |
| 2 | 0.5 | 99.3 |
| 0.5 | 0.8 | 97.9 |
| 1 | 0.8 | 97.9 |
| 2 | 0.8 | 98.9 |

## 6. CONCLUSION

We present an end-to-end heterogeneous face recognition solution for visible-to-thermal images using a novel Unit-Class Loss and a Cross-Modality Discriminator block. We initialize the backbone network with weights trained on a large-scale visible dataset. Next, the multimodal features are learned through training on a visible–thermal database using cross-entropy loss. The backbone network is then integrated with the proposed Unit-Class Loss and CMD block for HFR. The proposed loss function maximizes positive-to-negative pair distance by reducing intraclass variations while increasing interclass variance. The HFR performance is further enhanced by the addition of the CMD block to classify image pairs of the same or different classes. The experiments on multiple cross-modality face databases prove that the proposed method outperforms existing state-of-the-art methods on visible–thermal and visible–infrared datasets, validating the effectiveness of the proposed method.

The usage of the proposed loss and discriminator functions is simple and can be adapted to any network architecture for other HFR modalities. Furthermore, the methodology can be adapted for generic processing machines given its low computational complexity, making further research and industrial application feasible. In the future, more advanced architectures can be adapted to further improve and fine-tune the proposed strategy for other heterogeneous problems. Incorporating modality labels into training is also worth exploring. In the future, we will explore optimizing our network for other heterogeneous recognition problems.

Conflict of Interest:
The author(s) declare(s) that there is no conflict of interest regarding the publication of this article.

Funding Statement:
This work was supported by the Technology Innovation Program [grant number 20011756] funded By the Ministry of Trade, Industry & Energy (MOTIE, Korea).